\documentclass[10pt,twocolumn,letterpaper]{article}

\usepackage[pagenumbers]{cvpr} % To force page numbers, e.g. for an arXiv version

\usepackage{graphicx}
\usepackage{booktabs}
\usepackage{cite}
\usepackage{color}
\usepackage{xspace}
\usepackage{ctable}
\usepackage{colortbl}
\usepackage{subcaption}
\usepackage{multirow}
\usepackage{float}
\usepackage{stfloats}
\usepackage{stmaryrd}
\usepackage{algorithm}
\usepackage{algorithmicx}
\usepackage{algpseudocode}
\usepackage{threeparttable}
\usepackage{wrapfig}
\usepackage{caption}[font={small}]
\usepackage{autobreak}
\usepackage{amsmath}
\usepackage{amssymb}
\usepackage{bm}

\usepackage[pagebackref,breaklinks,colorlinks]{hyperref}

\usepackage[capitalize]{cleveref}
\crefname{section}{Sec.}{Secs.}
\Crefname{section}{Section}{Sections}
\Crefname{table}{Table}{Tables}
\crefname{table}{Tab.}{Tabs.}

\begin{document}

\title{FLAG3D: A 3D Fitness Activity Dataset with Language Instruction}

\author{
 Yansong Tang$^{*,\dagger,1}$, Jinpeng Liu$^{*,1}$, Aoyang Liu$^{*,1}$, \\ Bin Yang$^{1}$, Wenxun Dai$^{1}$, Yongming Rao$^{2}$, Jiwen Lu$^{\diamond, 2}$, Jie Zhou$^{2}$, Xiu Li$^{\diamond,1}$\\
 {\tt\small$^{*}$equal contribution, $^{\dagger}$project lead, $^{\diamond}$corresponding authors}\\
 \{$^{1}$Shenzhen International Graduate School, $^{2}$Department of Automation\}, Tsinghua University
 % Department of Automation, Tsinghua University\\
 }

\twocolumn[{
    \renewcommand\twocolumn[1][]{#1}
    \maketitle
    \begin{center}
        \includegraphics[width=1.0\linewidth]{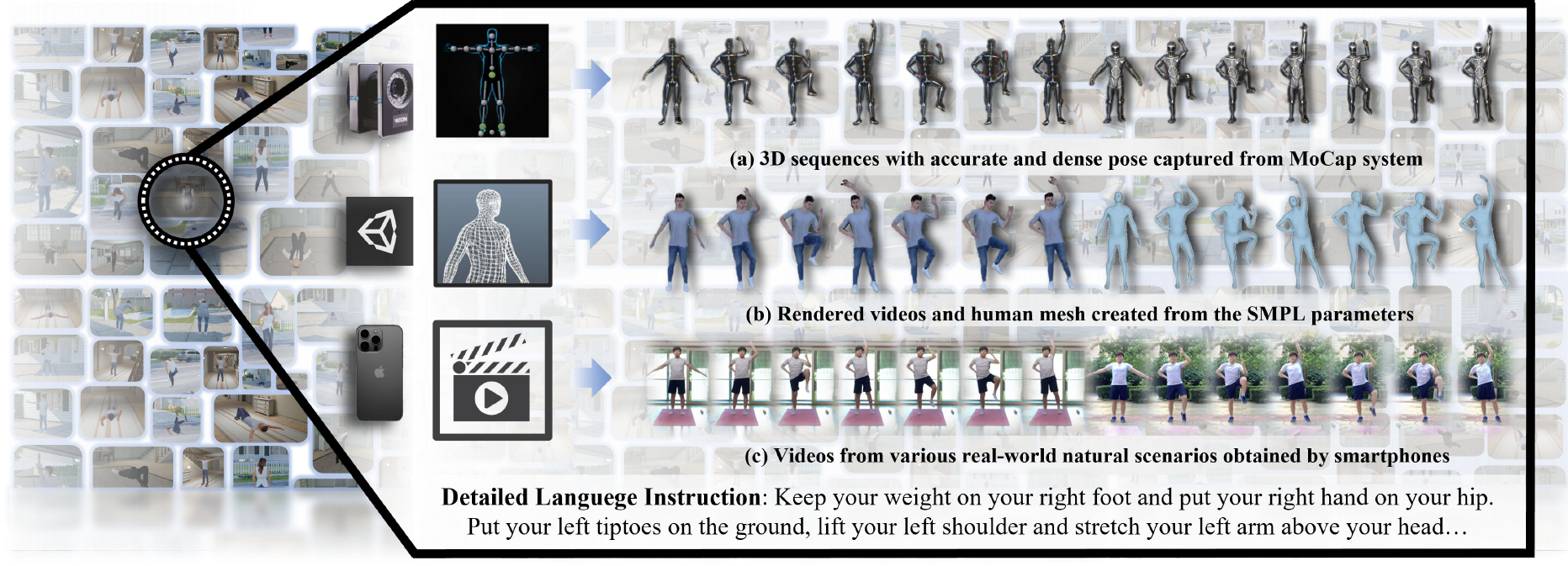}
        \captionof{figure}{An overview of the proposed FLAG3D dataset, which contains 180K videos of 60 daily fitness activities. Our dataset is comprised of (a) 3D activity sequences captured from advanced MoCap system, (b) rendered videos of different people with their SMPL parameters, and (c) real-world videos obtained by cost-effective phones from both indoor and outdoor natural environments. FLAG3D also provides a series of detailed and professional sentence-level language instructions for each fitness activity. All figures are best viewed in color.}
        \label{fig:teaser}
    \end{center}
}]

%%%%%%%%% BODY TEXT

\begin{abstract}
With the continuously thriving popularity around the world, fitness activity analytic has become an emerging research topic in computer vision. While a variety of new tasks and algorithms have been proposed recently, there are growing hunger for data resources involved in high-quality data, fine-grained labels, and diverse environments. In this paper, we present FLAG3D, a large-scale 3D fitness activity dataset with language instruction containing 180K sequences of 60 categories. FLAG3D features the following three aspects: 1) accurate and dense 3D human pose captured from advanced MoCap system to handle the complex activity and large movement, 2) detailed and professional language instruction to describe how to perform a specific activity, 3) versatile video resources from a high-tech MoCap system, rendering software, and cost-effective smartphones in natural environments. Extensive experiments and in-depth analysis show that FLAG3D contributes great research value for various challenges, such as cross-domain human action recognition, dynamic human mesh recovery, and language-guided human action generation. Our dataset and source code are publicly available at \url{https://andytang15.github.io/FLAG3D}.
\end{abstract}

\section{Introduction}
With the great demand of keeping healthy, reducing high pressure from working and staying in shape,
fitness activity has become more and more important and popular during the past decades~\cite{aifit}. 
According to the statistics\footnote{https://policyadvice.net/insurance/insights/fitness-industry-statistics}, there are over 200,000 fitness clubs and 170 million club members all over the world. 
More recently, because of the high expense of coaches and out-breaking of COVID-19, increasing people choose to exclude gym membership and do the workout by watching the fitness instructional videos from fitness apps or YouTube channels (\textit{e.g.,} FITAPP, ATHLEAN-X, The Fitness Marshall, \textit{etc.}).

Therefore, it is desirable to advance current intelligent vision systems to assist people to perceive, understand and analyze various fitness activities.

In recent years,
a variety of datasets have been proposed in the field~\cite{aifit,EC3D,verma2020yoga}, which have provided good benchmarks for preliminary research. 
However, these datasets might have limited capability to model complex poses, describe a fine-grained activity, and generalize to different scenarios. 
We present FLAG3D in this paper, a 3D \textbf{F}itness activity dataset with \textbf{LA}n\textbf{G}uage instruction. Figure~\ref{fig:teaser} presents an illustration of our dataset, which contains 180K sequences of 60 complex fitness activities obtained from versatile sources, including a high-tech MoCap system, professional rendering software, and cost-effective smartphones. In particular, FLAG3D advances current related datasets from the following three aspects:

\begin{table*}[t]
\caption{
\small Comparisons of FLAG3D with the relevant datasets. FLAG3D consists of 180K sequences (\texttt{Seqs}) of 60 fitness activity categories (\texttt{Cats}). It contains both low-level features, including 3D key points (\texttt{K3D}) and SMPL parameters, as well as high-level language annotation (\texttt{LA}) to instruct trainers, sharing merits of multiple resources from MoCap system in laboratory (\texttt{Lab}), synthetic (\texttt{Syn.}) data by rendering software and natural (\texttt{Nat.}) scenarios.
We evaluate various tasks in this paper, including human action recognition (\texttt{HAR}), human mesh recovery (\texttt{HMR}), and human action generation (\texttt{HAG}), while more potential applications like human pose estimation (\texttt{HPE}), repetitive action counting (\texttt{RAC}), action quality assessment and visual grounding could be explored in the future (see Section \ref{sec:application} for more details.).
}
\label{tab:actionsetcomparison}
\centering
\scriptsize
\vspace{2mm}
\resizebox{\textwidth}{!}{
\begin{tabular}{lccccccccr}
  \toprule
  
   Dataset & Subjs & Cats & Seqs & Frames    & LA & K3D & SMPL& Resource &  Task   \\
  
  \midrule
    
     PoseTrack~\cite{andriluka2018posetrack} & - & - & 550 & 66K    &$\times$ &$\times$ & $\times$ & Nat. & HPE\\  
  Human3.6M~\cite{ionescu2013human3} & 
  11 & 17 & 839 & 3.6M  & $\times$  & $\checkmark$ & - &  Lab & HAR,HPE,HMR 
    \\
      CMU Panoptic~\cite{joo2015panoptic} & 
  8 & 5 & 65 & 594K   & $\times$  & $\checkmark$ & - &Lab & HPE \\
      MPI-INF-3DHP~\cite{mehta2017monocular} & 
  8 & 8 & - & $>$1.3M    & $\times$  & $\checkmark$ & - &   Lab+Nat. & HPE,HMR \\
     3DPW~\cite{von2018recovering} &   7 & - & 60 & 51k &$\times$ & $\times$ & $\checkmark$ & Nat. &HMR  \\ 
    ZJU-MoCap~\cite{peng2021neural} & 
  6 & 6 & 9 & $>$1k & $\times$ & $\checkmark$ & $\checkmark$ &Lab & HAR,HMR \\
    NTU RGB+D 120~\cite{liu2019ntu} & 
  106 & 120 & 114k & -  & $\times$  & $\checkmark$ & - & Lab & HAR,HAG  \\
  
  HuMMan~ \cite{humman} & 1000 & 500 & 400K & 60M   & $\times$ & $\checkmark$ & $\checkmark$ & Lab & HAR,HMR\\

    \midrule
    HumanML3D~ \cite{HumanML3D}& - & - & 14K & -  &$\checkmark$ &$\checkmark$ & $\checkmark$ & Lab & HAG  \\
    KIT Motion Language~ \cite{MatthiasPlappert2016TheKM} & 111 & - & 3911 & -    &$\checkmark$ & $\checkmark$ & - & Lab & HAG  \\
    HumanAct12 ~\cite{action2motion}& 12 & 12 & 1191 & 90K   &$\times$ & $\times$ & $\checkmark$ & Lab & HAG  \\
    UESTC ~\cite{ji2019large} & 118 & 40 & 25K &  $>$ 5M & $\times$ & $\checkmark$ & - &  Lab & HAR,HAG\\
    \midrule
     Fit3D ~\cite{aifit}& 13 & 37 & - & $>$ 3M  & $\times$  & $\checkmark$ & $\checkmark$ &Lab &HPE,RAC \\
     EC3D ~\cite{EC3D}& 4 & 3 & 362 & -   &$\times$  & $\checkmark$ & - & Lab &HAR \\
     Yoga-82 ~\cite{verma2020yoga}& - & 82 & - & 29K   & $\times$ & $\times$ & $\times$ &Nat. & HAR,HPE \\
  \midrule
  
  \textbf{FLAG3D (Ours)}  & 10+10+4 & 60 & 180K & 20M   & $\checkmark$ &$\checkmark$  & $\checkmark$ & Lab+Syn.+Nat. & HAR,HMR,HAG \\
  \bottomrule
  \end{tabular}
}

\end{table*}

\textbf{Highly Accurate and Dense 3D Pose.} For fitness activity, there are various poses within lying, crouching, rolling up, jumping \textit{etc.}, which involve heavy self-occlusion and large movements. These complex cases bring inevitable obstacles for conventional appearance-based or depth-based sensors to capture the accurate 3D pose. To address this, we set up an advanced MoCap system with 24 VICON cameras~\cite{vicon} and professional MoCap clothes with 77 motion markers to capture the trainers’ detailed and dense 3D pose.

\textbf{Detailed Language Instruction.} Most existing fitness activity datasets merely provide a single action label or phase for each action~\cite{EC3D,verma2020yoga}. However, understanding fitness activities usually requires more detailed descriptions. We collect a series of sentence-level language instructions for describing each fine-grained movement. Introducing language would also facilitate various research regarding emerging multi-modal applications.

\textbf{Diverse Video Resources.}
To advance the research directly into a more general field, we collect versatile videos for FLAG3D. Besides the data captured from the expensive MoCap system, we further provide the synthetic sequences with high realism produced by rendering software and the corresponding SMPL parameters. In addition, FLAG3D also contains videos from natural real-world environments obtained by cost-effective and user-friendly smartphones.

To understand the new challenges in FLAG3D, 
we evaluate a series of recent advanced approaches and set a benchmark for various tasks, including skeleton-based action recognition, human mesh recovery, and dynamic action generation. 
Through the experiments, we find that 
1) while the state-of-the-art skeleton-based action recognition methods have attained promising performance with highly accurate MoCap data under the in-domain scenario, the results drop significantly under the out-domain scenario regarding the rendered and natural videos.
2) Current 3D pose and shape estimation approaches easily fail on some poses, such as kneeling and lying, owing to the self-occlusion. FLAG3D provides accurate ground truth for these situations, which could improve current methods’ performance in addressing challenging postures.
3) Motions generated by state-of-the-art methods appear to be visually plausible and context-aware at the beginning. However, they cannot follow the text description faithfully as time goes on.

To summarize, our contributions are twofold: 1) We present a new dataset named FLAG3D with highly accurate and dense 3D poses, detailed language instruction, and diverse video resources, which could be used for multiple tasks for fitness activity applications. 2) We conduct various empirical studies and in-depth analyses of the proposed FLAG3D, which sheds light on the future research of activity understanding to devote more attention to the generality and the interaction with natural language.

\section{Related Work}
Table~\ref{tab:actionsetcomparison} presents a comparison of our FLAG3D with the related datasets. FLAG3D provides detailed language instructions for text-driven action generation compared with other datasets. Here we briefly review numbers of relevant datasets and methods regarding the three tasks we focus on.

\noindent \textbf{Human Action Recognition.}
As the foundation of video understanding, pursuing diverse datasets has never stopped in action recognition. Existing works have explored various modalities in this area, such as RGB videos \cite{kuehne2011hmdb,soomro2012ucf101,carreira2017quo,shao2020finegym}, optical flows \cite{simonyan2014two}, audio waves \cite{xiao2020audiovisual}, and skeletons \cite{yan2018spatial}. Among these modalities, skeleton data draws increasing attention because of its robustness to environmental noises and action-focusing nature. During the past few years, various network architectures have been exploited to model the spatio-temporal evolution of the skeleton sequences, such as different variants of RNNs \cite{du2015hierarchical, song2017end, zhang2017view}, CNNs \cite{choutas2018potion, yan2019pa3d, liu2017two, duan2022revisiting} and GCNs\cite{yan2018spatial, shi2019two, li2019actional, liu2020disentangling, chen2021channel, tang2018deep}. In terms of skeletal keypoint, current action recognition datasets can be divided into two classes: one is 2D keypoint datasets \cite{yan2018spatial, duan2022revisiting} extracted by pose estimation methods \cite{cao2017realtime, chen2018cascaded, martinez2017simple, sun2019deep}, and the other is 3D keypoint datasets \cite{shahroudy2016ntu, liu2019ntu, ionescu2013human3, omran2018neural} collected by sensors or other sophisticated equipment. However, most existing datasets are limited to a single domain of natural scenes. FLAG3D dataset takes a different step towards cross-domain action recognition between rendered videos and real-world scenario videos.

\noindent \textbf{Human Mesh Recovery.}
Human mesh recovery obtains well-aligned and physically plausible mesh results that human models can parametrize, such as SMPL\cite{loper2015smpl}, SMPL-X\cite{pavlakos2019expressive}, STAR\cite{osman2020star} and GHUM\cite{xu2020ghum}. Current methods take keypoints\cite{bogo2016keep, pavlakos2019expressive, Zhang2021LightWeight}, images\cite{gu2018ava, guler2019holopose, kocabas2021pare, kolotouros2019learning, li2020hybrik, omran2018neural, pavlakos2018learning}, videos\cite{choi2021beyond, kanazawa2019learning, luo20203d, mehta2020xnect, moon2020i2l, sun2019human} and point clouds\cite{bhatnagar2020combining, hong2021garmentd, jiang2019skeleton, liu2021votehmr, wang2021locally} as inputs to recover the parametric human model under optimization\cite{bogo2016keep, zanfir2018monocular, lassner2017unite} or regression\cite{kanazawa2018end, pavlakos2018learning, omran2018neural, tung2017self, kolotouros2019learning} paradigm. Besides the above input modalities, there are ground-truth SMPL parameters provided by human datasets. They are registered by marker-less multi-view MoCap\cite{peng2021neural, Yu2020HUMBIAL, mehta2017monocular, pavlakos2019expressive, Mehta2018SingleShotM3, Zhang2020ObjectOccludedHS}, or marker/sensor based Mocap\cite{Sigal2009HumanEvaSV, ionescu2013human3, von2018recovering}. SMPL can also be fitted with the rendered human scan in synthesis datasets\cite{Varol2017LearningFS, cai2021playing, patel2021agora, tao2021function4d}. Easily recovered human poses of existing datasets cause the performance of human mesh recovery algorithms not to be fairly evaluated\cite{pang2022benchmarking}, whereas FLAG3D provides human poses with heavy self-occlusion and large movements. The work most related to ours is HuMMan~\cite{humman}, which contains large-scale and comprehensive multi-modal resources captured in a single MoCap room. In comparison, FLAG3D is complementary with rendered and natural videos, as well as more detailed language instructions to describe the activity. 

\noindent \textbf{Human Action Generation.}
In the past several years, various works have utilized multiple forms of information to guide the generation of human actions. Among these, one direction \cite{HaoyeCai2017DeepVG, ZhenyiWang2019LearningDS, PingYu2020StructureAwareHG, ChuanGuo2022Action2videoGV, action2motion, MathisPetrovich2021ActionConditioned3H} is to explore the underlying data structure of action sequences based on action categories. As real-life movements are often accompanied by audio messages, another direction\cite{KentaTakeuchi2017SpeechtoGestureGA, EliShlizerman2017AudioTB, TaoranTang2018DanceWM, HsinYingLee2019DancingTM, RuoziHuang2020DanceRL} is to use the audio and motion timing alignment feature. Translating text descriptions to human motion is an emerging topic as well. Several works \cite{ChaitanyaAhuja2019Language2PoseNL, MatthiasPlappert2018LearningAB, TatsuroYamada2018PairedRA, AninditaGhosh2021SynthesisOC, petrovich22temos, HumanML3D, GuyTevet2022MotionCLIPEH, AlecRadford2021LearningTV, zhang2022motiondiffuse, tevet2022human, song2020denoising, ho2020denoising} try to match semantic information and high-dimensional features of action sequences so that it could pursue natural motion sequences guided by language. For traditional motion generation tasks, HumanAct12\cite{action2motion}, UESTC\cite{YanliJi2018ALR} and NTU RGB+D\cite{liu2019ntu} are three commonly used benchmarks. However, the above datasets do not provide paired sophisticated semantic labels to the motion sequences. Moreover, there are not enough motions for the exact text in the language-motion dataset KIT\cite{MatthiasPlappert2016TheKM}. Recently, BABEL\cite{AbhinandaRPunnakkal2021BABELBA} and HumanML3D\cite{HumanML3D} re-annotates AMASS\cite{NaureenMahmood2019AMASSAO} with English language labels. Nonetheless, they focus on simple actions with uncomplicated descriptions. FLAG3D provides long sequences of actions with detailed and professional language instructions.
\begin{figure*}[!t]
\begin{center}
   \includegraphics[width=1.0\linewidth]{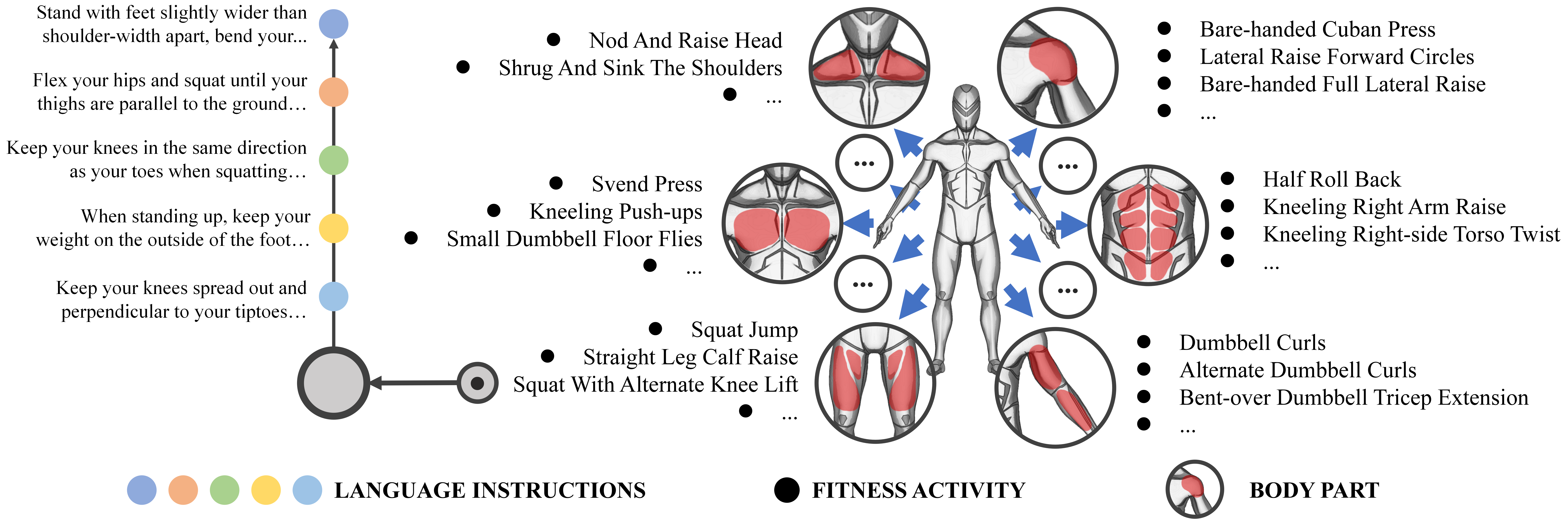}
\end{center}
\caption{An illustration of the taxonomy of our FLAG3D dataset, which is systematically organized in three levels as \textit{body part}, \textit{fitness activity} and \textit{language instruction}. This figure details a concrete example of the \textit{``Squat With Alternate Knee Lift''} activity that is mainly driven by the quadriceps femoris muscle of the \textit{``Leg''}, while the corresponding language instructions are shown in the left.
}
\label{fig:lexicon}
\end{figure*}

\section{The FLAG3D Dataset}
\label{sec:dataset}
\subsection{Taxonomy}
The first challenge to construct FLAG3D is establishing a systematic taxonomy to organize various fitness activities. In previous literature, most existing fitness datasets~\cite{aifit,EC3D,verma2020yoga} mix up all the activities. We present a deeper hierarchical lexicon as shown in Figure~\ref{fig:lexicon}, which contains three levels from roots to leaves, including body parts, fitness activity, and language instructions.

(1) \textit{Body Part.} 
For the first level, we share our thoughts with HuMMan~\cite{humman} which uses the driving muscles as basic categories. However, numerous fine-grained muscles exist in the human body, and one activity might be driven by different muscles. We follow the suggestions of our fitness training coaches and choose ten parts of the human body with rich muscles as \textit{chest, back, shoulder, arm, neck, abdomen, waist, hip, leg} and \textit{multiple parts}\footnote{Some activities are driven by muscles of various body parts.}.

(2) \textit{Fitness Activity.} Sixty everyday fitness activities are selected for the second level, linked to the corresponding body parts of the first level. For example, the activity \textit{``Squat With Alternate Knee Lift’’} is associated with the quadriceps femoris muscle of the \textit{``Leg’’}. We include the complete list of the 60 fitness activities in the Appendix.

(3) \textit{Language Instruction.} We compose the third level of the lexicon with a set of language descriptions from the guidance of the training coaches to instruct users to accomplish the fitness activity. As an example shown in Figure~\ref{fig:lexicon}, the fitness activity \textit{``Squat With Alternate Knee Lift’’} is detailed as \textit{``Stand with feet slightly wider than shoulder-width apart, bend your elbows and put your hands in front of your chest. Flex your hips and squat until your thighs are parallel to the ground, and keep your knees in the same direction as your toes when squatting...’’}. There are about 3 sentences and 57 words for each fitness activity on average.

\subsection{Data Collection}
We deploy high-precision MoCap equipment in an open lab to capture accurate human motion information. To obtain the rendered videos, we purchase kinds of virtual scenes and character models to make full use of the collected 3D skeleton sequences. In different environments, we record real-world natural videos. We agree with the volunteers and ensure that researchers can use these data. More details can be found in the supplementary materials. We detail the data collection process below.

\noindent \textbf{Data from MoCap System.}
Our MoCap system is equipped in a lab of 20 meters long, 8 meters wide, and 7 meters high. The lab uses the high-tech VICON~\cite{vicon} MoCap system to capture the actors’ body part movements through optical motion capture. Cameras used in this system have a maximum resolution of 4096$\times$4096. It is capable of 120fps while maintaining maximum resolution sampling. Ten volunteers perform the actions in the motion capture field. Infrared cameras transmit the high frame rate IR gray-scale images captured over the fiber optic cable to the data switch. They are clock aligned and finally sent to a device with specialized processing software. Through the professional machine, we can monitor the movements of volunteers in various forms, including masks, bones, and marker points. Moreover, we hire professional technicians to perform data restoration and motion  retargeting based on high-precision original data so that we can ensure the accuracy and diversity of provided 3D motion data. Meanwhile, we ask each performer to wear MoCap clothes with 77 motion markers listed on a Table in supplementary materials. Dense marker points are also a safeguard for our 3D motion data. Before performing the activity, we ask them to watch the instructional video and read the language instructions. For each action, eight males and two females will perform three times, each containing over eight repetitions. In total, we have 7200 motion sequences, where
$
7200 = 10 (\textit{people}) \times 3 (\textit{times}) \times 60 (\textit{actions}) \times 4 (\textit{motion retargeting})
$.

\begin{figure*}[!t]
\begin{center}
   \includegraphics[width=0.9\linewidth]{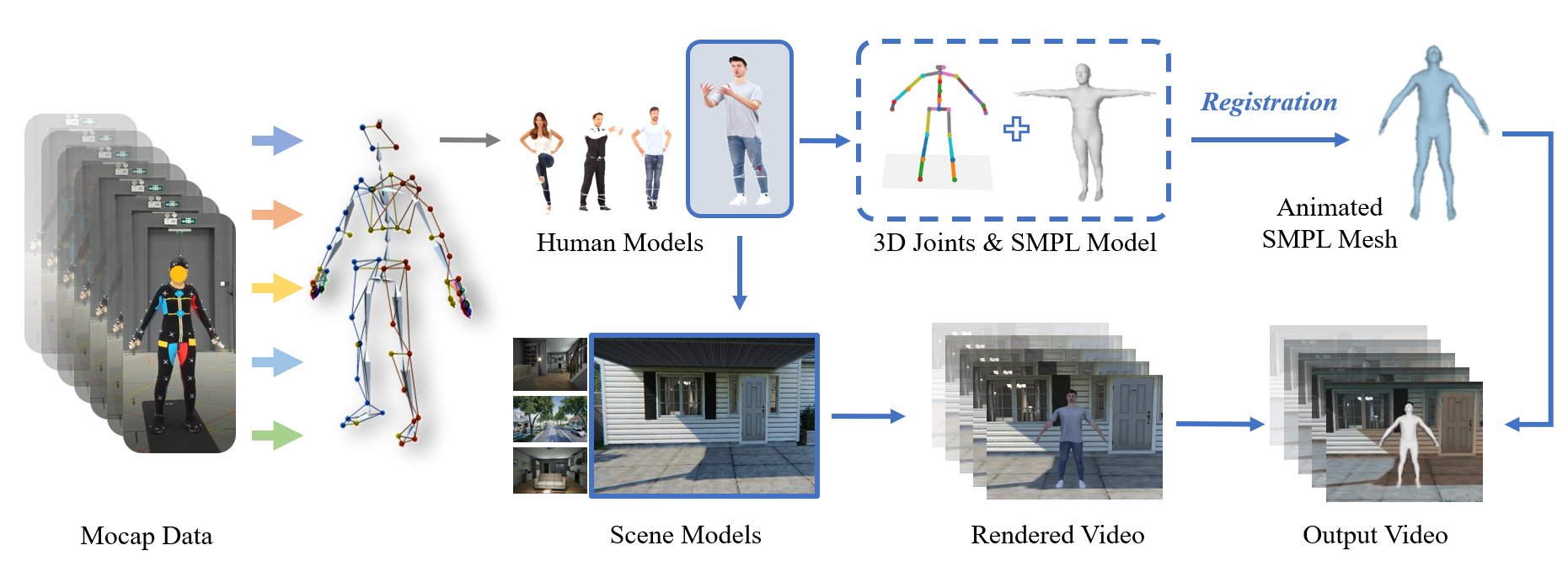}
\end{center}

\caption{
A process of producing the rendered videos and the SMPL parameters. First, we apply the MoCap data to get the dynamic human poses in virtual scenes and render RGB videos with camera parameters. Then we use this joint information from dynamic human models to recover the human mesh in SMPL format. Finally, we combined the SMPL mesh and RGB videos to display the output.
}
\label{fig:SMPL_recovery}
\end{figure*}

\noindent \textbf{Data from Rendering Software.}
To fully utilize the 3D MoCap data, we use the rendering software Unity3D~\cite{Unity3D} to produce synthetic 2D videos with RGB color. For 2D videos, we purchase realistic scene models in Unity Asset Store~\cite{Unity_Asset}, including indoor and outdoor scenes. As well, we select 6 camera positions in each scene. Our camera positions are dispersed around the avatar. However, we change parameters such as the focal length of each camera to ensure that the viewfinder fits and that the camera parameters are diverse. Specifically, we are greatly appreciated that Renderpeople\cite{Renderpeople} provides several free character models. We select 4 avatars, import the skeleton information into these avatars and record the motion of the avatars in all directions. The resolution of these videos is 854 × 480, and the frame rate is 30fps. Totally, we have 172,800 videos, where
$
172,800 = 1800 (\textit{mocap sequence}) \times 6 (\textit{camera position}) \times 4 (\textit{avatar}) \times 4 (\textit{virtual scences})
$.

\noindent \textbf{Data from Real-world Environment.}
To obtain versatile data resources and add diverse scenes, we ask 10 extra people to record videos in different real-world scenarios. The recording process is executed using smartphones that can capture 1080p videos from the front view and side view simultaneously, to ensure the diversity of shooting angles. As well, before performing the activity, volunteers are asked to watch the instructional video and read the language instructions carefully. We have 7200 videos, where 
$
7200 = 10 (\textit{people}) \times 3 (\textit{times}) \times 60 (\textit{actions}) \times 2 (\textit{views}) \times 2 (\textit{scenes})
$.

Therefore, we have 24 subjects in all of the videos, where $24=10(\textit{mocap})+10(\textit{real-world})+4(\textit{render-people})$.

\subsection{The Body Model}

To facilitate different applications (\textit{e.g.,} human mesh recovery and human action generation), FLAG3D adopts the SMPL\cite{loper2015smpl} parametric model because of its ubiquity and generality in various downstream tasks.

Specifically, the SMPL parameters comprise pose parameters $\theta \in \mathbb{R}^{N\times 72}$, shape parameters $\beta \in \mathbb{R}^{N \times 10}$ and translation parameters $t \in \mathbb{R}^{N \times 3}$, where $N$ is the number of frames for each video.

We obtain the SMPL parameters based on the captured keypoints and an optimization algorithm. In particular, the optimization process is composed of two stages, where the first stage is to get the shape parameter $\beta \in \mathbb{R}^{N \times 10}$, and the second stage is to gain the pose $\theta \in \mathbb{R}^{N\times 72}$ as well as translation parameter $t \in \mathbb{R}^{N \times 3}$.
We denote the $E_s$ and $E_p$ as two objective functions for shape and pose optimization. In the first stage, the objective function is formulated as follows:
\begin{align}
    E_s(\beta)=&\frac{\lambda_1}{N}\sum_{(i,j)\in \mathcal{L}}||\bm{J}_i(\mathbb{M}(\beta))-\bm{J}_j(\mathbb{M}(\beta))-\mathcal{P}(\bm{g}_i-\bm{g}_j)||_2^2
    \nonumber
    \\
    &+
    \lambda_{2} ||\beta||_2^2.
    \nonumber
\end{align}

Here $J_i$ is the joint regressor for joint $i$, $\bm{g}$ is the ground truth skeleton, and $\mathbb{M}$ is the parametric model\cite{loper2015smpl}. $\mathcal{L}$ and $\mathcal{J}$ represent the body limbs and joint sets, respectively. $\mathcal{P}$ is the projection function that projects the $g_i-g_j$ in the direction of $J_i-J_j$.
Similarly, the objective function in the second stage is as follows:
\begin{equation}
    E_p(\theta,t)=\lambda_3\frac{1}{N}\sum_{j \in \mathcal{J}}\lambda_{p1}||\bm{J}_j(\mathbb{M}(\theta,t))-\bm{g}_j||_2^2+\lambda_4||\theta||_2^2.
    \nonumber
\end{equation}

In the equations above, different weights of $\lambda_k (k=1,2,3,...) $ are denoted for each loss term (see supplementary materials for details). We adopt the L-BFGS\cite{liu1989limited} method where the search step-length satisfies the strong Wolfe conditions\cite{wolfe1969convergence} for solving this optimization problem because of its memory and time efficiency.

\section{Experiments}

\subsection{Human Action Recognition}
 FLAG3D contains both RGB videos and 3D skeleton sequences, maintaining abundant resources for 2D and 3D skeleton-based action recognition. Moreover, FLAG3D also provides videos from different domains, allowing us to evaluate different models’ generalized abilities. We first report human action recognition accuracy with the 3D skeleton data captured from the MoCap system. Then we use the 2D skeleton data extracted from both rendered and real-world videos to test the transferable ability of the models.

 \noindent \textbf{Experiment Setup.}
For the in-domain evaluation, we use the 5040 skeleton sequences of 7 subjects for training, while the other 2160 skeleton sequences of 3 subjects for testing. For the cross-domain evaluation, we follow \cite{duan2022revisiting} to use the Top-Down approach for 2D pose extraction in both rendered and real-world videos. For data selection, we select the rendered videos from the front and side view in one scene (7200 $\times$ 3) for training samples and take all 7200 real-world videos for testing. We evaluate five state-of-the-art methods as ST-GCN\cite{yan2018spatial}, 2s-AGCN\cite{shi2019two}, MS-G3D\cite{liu2020disentangling}, CTR-GCN\cite{chen2021channel} and PoseC3D\cite{duan2022revisiting}. We exclude PoseC3D from in-domain experiments since it only supports 2D keypoint input. Table \ref{tab:action} presents the compared results. We also test our model using the Mindspore \cite{MindSpore}.

\begin{table}[t]
\centering
\setlength{\tabcolsep}{15pt}
\caption{Action recognition accuracy on the FLAG3D dataset.}
\label{tab:action}
\begin{tabular}{lcc}
    \hline
    Method & In-domain & Out-domain \\  
    \toprule
    ST-GCN~\cite{yan2018spatial} & 97.8 & 69.9 \\
    2s-AGCN~\cite{shi2019two} & 98.6 & 81.6\\
    MS-G3D~\cite{liu2020disentangling} & 97.7 & 73.6 \\
    CTR-GCN~\cite{chen2021channel} & 97.5 & 77.2 \\
    PoseC3D~\cite{duan2022revisiting} & - & 79.9 \\
    \bottomrule    
\end{tabular}
\end{table}

\begin{table}[t]
\centering
\caption{Results of transfer learning on FineGym and NTU60.}
\label{tab:transfer}
\begin{tabular}{c|c|c}
    \hline
    Method & FineGym &  NTU60-XSub \\  
    \toprule
    ST-GCN~ & 91.4 / \textbf{92.0} (\textbf{+0.6}) & 89.0 /  \textbf{89.0} (\textbf{+0.0}) \\
    2s-AGCN~ & 91.8 / \textbf{92.1} (\textbf{+0.3}) & 89.7 / \textbf{91.0} (\textbf{+1.3}) \\
    MS-G3D~ & 92.7 / \textbf{93.4} (\textbf{+0.7})& 92.2 / \textbf{92.3} (\textbf{+0.1})\\
    CTR-GCN~ & 92.9 / \textbf{93.5} (\textbf{+0.6}) & 90.6 / \textbf{90.8} (\textbf{+0.2}) \\
    PoseC3D~ & 95.4 / \textbf{95.8} (\textbf{+0.4}) & 93.7 / \textbf{93.9} (\textbf{+0.2}) \\
    \bottomrule    
\end{tabular}
\end{table}

\begin{figure}[t]
\begin{center}
\includegraphics[width=\linewidth]{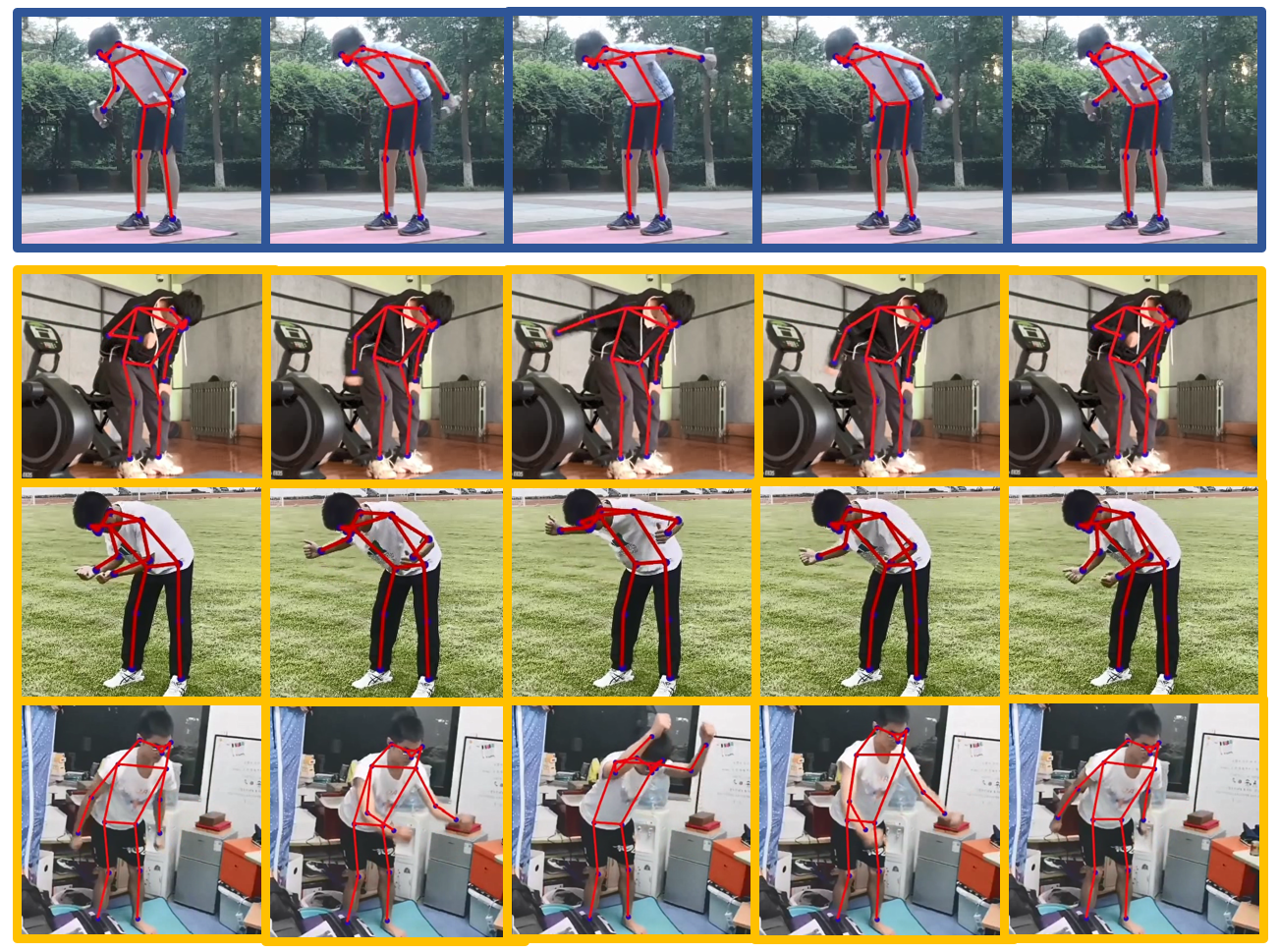}
\end{center}
\vspace{-15pt}
\caption{Case study of 2s-AGCN prediction results. The blue boxes are the selected frames of the target category, and the yellow boxes are the confusing categories. From top to bottom are \textit{``Bent-over Dumbbell Tricep Extension''}, \textit{``Right-side Bent-over Tricep Extension With Resistance Band''}, \textit{``Bent-over W-shape Stretch''} and \textit{``Bent-over Y-shape  Stretch''}.}
\label{fig:cf_cases}
\end{figure}

\noindent \textbf{Result and Analysis.}
For the in-domain experiments, the Top-1 accuracy of all models is high, which shows that our 3D skeleton data is effective with recently advanced algorithms. Regarding the out-domain experiments, the accuracy drops drastically when transferring the models from rendered to real-world scenarios. On the widely used NTU RGB+D 60\cite{shahroudy2016ntu} and 120 benchmark\cite{liu2019ntu}, Top-1 accuracy achieves and 96.6\% and 89.6\% respectively with PoseC3D\cite{duan2022revisiting}, but 79.9\% only on FLAG3D. Unlike NTU RGB+D, which has a large proportion of daily actions in the indoor environment, FLAG3D focuses more on the classification of fitness actions, which requires more attention to fine-grained action distinctions. Figure \ref{fig:cf_cases} shows these actions share sufficient similarities in motion patterns, such as bending over and swinging arms. As for \textit{``Lying Shoulder Joint Downward Round’’}, the counterpart - \textit{``Lying Shoulder Joint Upward Round’’} challenges the model in the aspect of temporal modeling. These categories require the models to focus on fine-grained action differences. The FLAG3D dataset can be served as a new benchmark for out-domain and fine-grained action understanding. Moreover, we finetune the models (pre-trained on FLAG3D) on FineGym and NTU60. In Table \ref{tab:transfer}, pre-trained models achieved better performance (\textbf{in bold}), especially on FineGym, which shares some common grounds with FLAG3D such as the fine-grained nature of sports. Promising results show that our FLAG3D dataset can transfer beneficial signals for pre-trained models to boost the performance of other datasets.

\begin{figure*}[!t]
\begin{center}
   \includegraphics[width=0.95\linewidth]{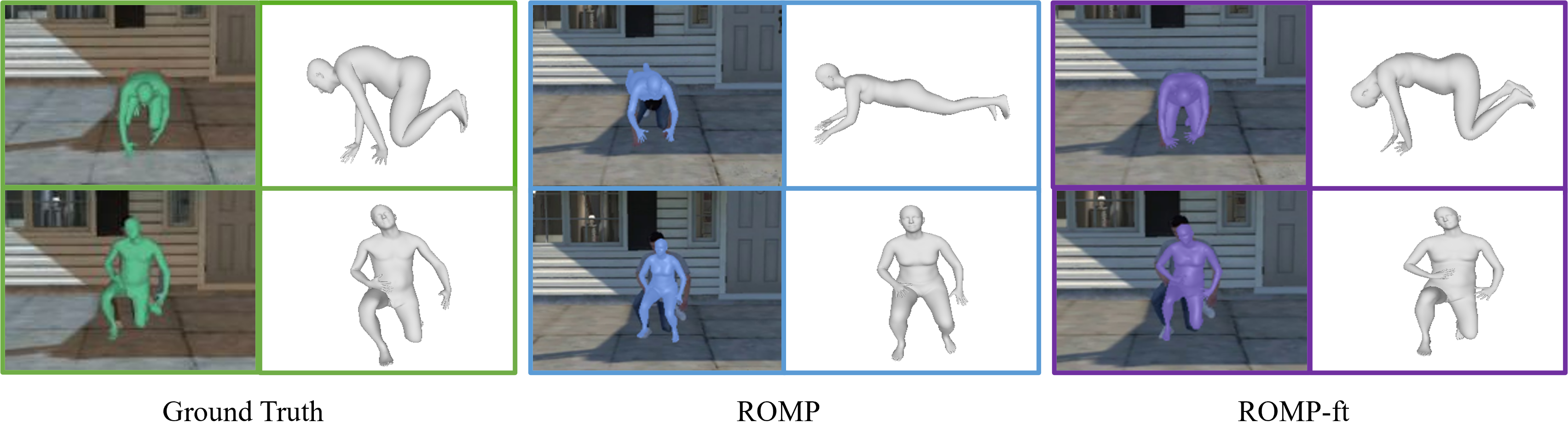}
\end{center}
\vspace{-6pt}
\caption{
Examples of SMPL prediction results. 
The top activity is prostrating and the bottom one is taking the knee. ROMP-ft ( \textit{i.e.,} fine-tuning ROMP on the training set of FLAG3D ) improves its ability to correctly estimate complex postures like kneeling.
}
\label{fig:SMPL_Results}
\end{figure*}

\subsection{Human Mesh Recovery}
FLAG3D provides the SMPL\cite{loper2015smpl} annotations, which are the prevalent ground truth in human mesh recovery. It is available to perform and evaluate popular methods for estimating 3D human poses and shapes. In this section, we first evaluate deep learning-based regression algorithms to verify that our dataset is qualified as a benchmark. Then we use the SMPL\cite{loper2015smpl} annotation data to train ROMP\cite{romp} to improve its performance on our test set. 

\noindent \textbf{Experiment Setup.}
To ensure diversity in the subset, we opt for 300K frames for each scene and view during data selection. In order to avoid potential continuity issues and information leakage (\textit{e.g.}, two videos with the same action and human model but different repetitions are in different datasets), we select the first 20\% videos for each scene and view them as the test set. We benchmark three typical methods: VIBE\cite{VIBE}, BEV\cite{BEV}, and ROMP\cite{romp} on FLAG3D using MPJPE (mean per joint position error) and PA-MPJPE (Procrustes-aligned mean per joint position error) metrics. Results are presented in Table \ref{tab:hmr}.

\begin{table}[t]
\centering
\caption{Human mesh recovery accuracy on the FLAG3D dataset. ``$\downarrow$" indicates that the lower value is better. ``ft" represents that we have fine-tuned this method on our trainset.}  
\label{tab:hmr}
\setlength{\tabcolsep}{15pt}
\begin{tabular}{lccc}
    \hline
    Method & MPJPE $\downarrow$  & PA-MPJPE $\downarrow$ \\  
    \toprule
    VIBE\cite{VIBE} & 376.67 & 106.27\\
    BEV\cite{BEV} & 382.77 & 117.62 \\
    ROMP\cite{romp} & 379.44 & 100.48\\
    \midrule
    ROMP-ft\cite{romp} &  \textbf{114.73} & \textbf{62.29}                    \\
    \bottomrule    
\end{tabular}
\end{table}

\begin{table}[hptb]
\small
\centering
\caption{Performance on challenging cases using two protocols.}
\label{tab:rb-hmr-2}
\setlength{\tabcolsep}{4pt}{
\begin{tabular}{l|cc|cc}
    \hline
  & \multicolumn{2}{|c|}{P1} & \multicolumn{2}{|c}{P2} \\
    \hline
           & MPJPE   & PA-MPJPE & MPJPE   & PA-MPJPE  \\  
    \hline
    w/o  & 260.918 &132.574 &490.248 & 111.654 \\
    w. ft-FLAG3D &\textbf{119.109} &  \textbf{81.179} &\textbf{131.001} & \textbf{75.428} \\
    \hline    
\end{tabular}}
\end{table}

\noindent \textbf{Result and Analysis.}
These methods without training achieved unsatisfactory MPJPE and PAMPJPE on the dataset. One of the most important reasons is that when the person in the rendered video is kneeling or lying, the task could be challenging because of the occlusion in the visual view. As displayed in Figure \ref{fig:SMPL_Results}, ROMP\cite{romp} interprets kneeling as lying and interprets taking on the knee as squatting on the ground. For the evaluation part, VIBE\cite{VIBE} and ROMP\cite{romp} achieved the best MPJPE and PA-MPJPE, respectively. But these metrics are still high, indicating that there is still much room for improvement of 3D shape estimation methods on the FLAG3D dataset. Therefore, FLAG3D can be served as a new benchmark for 3D pose and shape estimation tasks. Since ROMP\cite{romp} achieved the top-1 PA-MPJPE, we fine-tuned it on FLAG3D with HR-Net\cite{sun2019deep} backbone. ROMP\cite{romp} could handle challenging cases and reach better MPJPE and PA-MPJPE after being fine-tuned on our dataset. It indicates that our dataset could benefit 3D pose estimation approach to improve their performance. To verify our ideas, we also test videos involving challenging actions in protocol 1 and challenging views in protocol 2 as shown in table \ref{tab:rb-hmr-2}. Both situations with self-occlusion can be mitigated after fine-tuning on FLAG3D. 

\begin{table}[t]
\centering
\setlength{\tabcolsep}{15pt}
\caption{ Results of MDM in KIT dataset.}
\label{tab:kit}
\begin{tabular}{lcc}
    \hline
     & R-Precision $\uparrow$ & FID $\downarrow$  \\  
    \toprule
    w/o FLAG3D~ & 0.396 & 0.497  \\
    w. FLAG3D~ & \textbf{0.407}  & \textbf{0.491} \\
    \bottomrule    
\end{tabular}
\end{table}

\linespread{1.15}
\begin{table*}[t]
\setlength{\tabcolsep}{3pt}
\centering
\caption{
Results of human action generation. 
Multi.: MultiModality. APE: Average Position Error. AVE: Average Variance Error. }
\label{tab:hag}
\begin{tabular}{lccc|lcc|lccc}
    \hline
    \toprule
     \multicolumn{1}{l}{Method} & FID $\downarrow$  & Acc. $\uparrow$ & Multi.$\uparrow$ & \multicolumn{1}{l}{Method} & $\text{APE}_\text{root}$ $\downarrow$ & $\text{AVE}_\text{root}$ $\downarrow$ & \multicolumn{1}{l}{Method} & FID $\downarrow$ & R-precision$\uparrow$ & Multi.$\uparrow$\\     
    \hline
     \multicolumn{1}{l}{ACTOR\cite{MathisPetrovich2021ActionConditioned3H}} & \multicolumn{1}{l}{14.77}  & 94.50 & 6.53 & TEMOS\cite{petrovich22temos} & 0.61 & 0.66      
     & Guo \textit{et al.} \cite{HumanML3D} & 15.12 & 0.10  &  1.20 \\ 
    \bottomrule    
\end{tabular}
\end{table*}
\linespread{1}

\subsection{Human Action Generation}
Detailed language instructions and 3D skeleton-based motion sequences with SMPL\cite{loper2015smpl} annotations are included in FLAG3D, which facilitate the application of human action generation. This section reports the results of action-conditioned 3D human motion synthesis under both category-based settings and language-based settings. 

\noindent \textbf{Experiment Setup.}
For category-based action generation, we test the results in ACTOR\cite{MathisPetrovich2021ActionConditioned3H}. We first selected skeleton sequences of 5 subjects for training, while the other 5 subjects were for testing. Same as UESTC~\cite{YanliJi2018ALR} in ACTOR~\cite{MathisPetrovich2021ActionConditioned3H}, we use ST-GCN~\cite{yan2018spatial} as the feature extractor. For language-based action generation, we evaluate the methods of Guo \textit{et al.}\cite{HumanML3D} and TEMOS\cite{petrovich22temos}. We use 90\% language-motion pairs for training and 10\% for testing. 

\begin{figure}[t]
\begin{center}
   \includegraphics[width=1\linewidth]{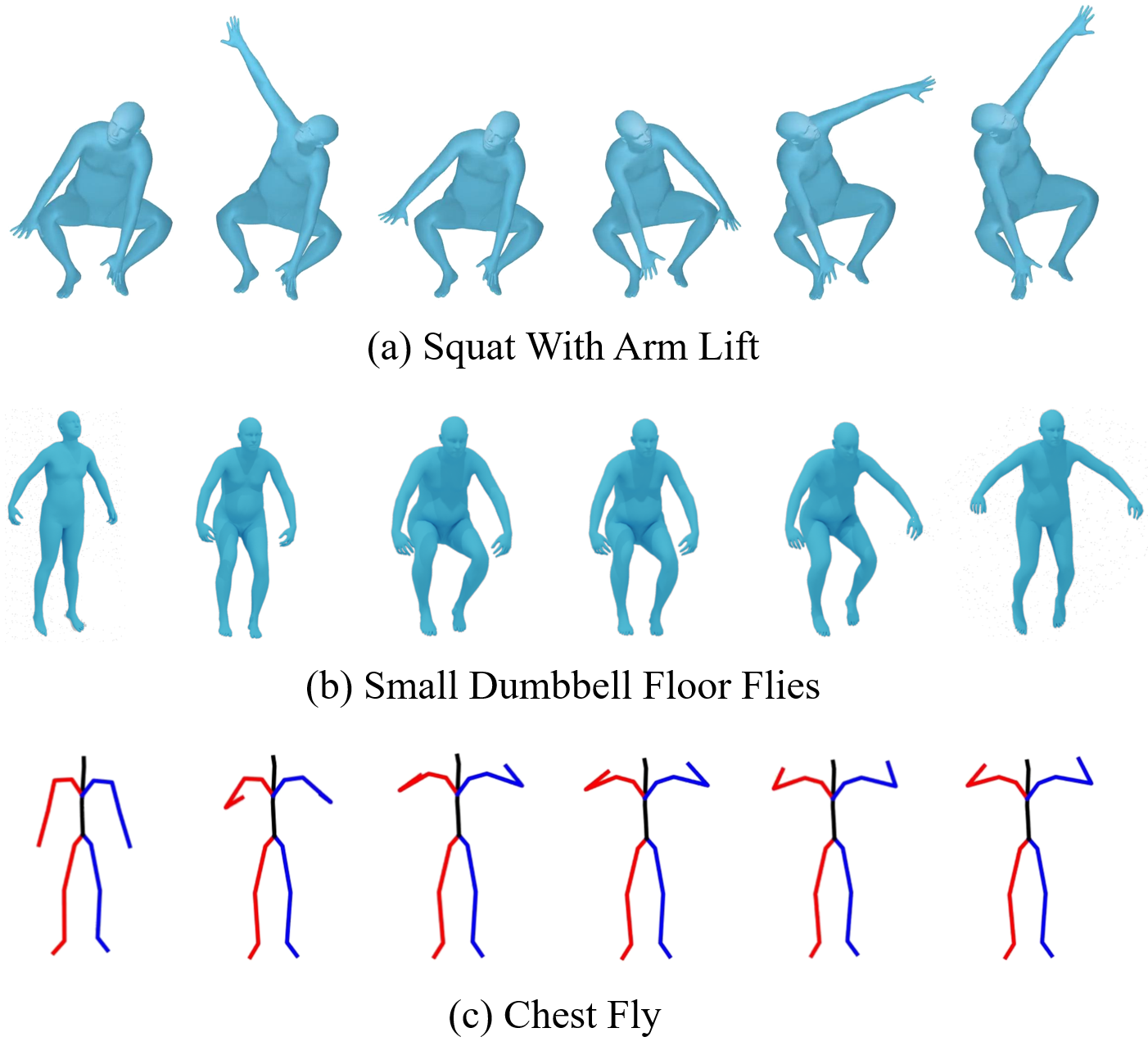}
\end{center}
\caption{Qualitative results in FLAG3D. \textbf{(a)} ``Squat With Arm Lift" visualization results in the category-based method. \textbf{(b)} Small Dumbbell Floor Flies: \textit{``Lie flat on your yoga mat, bend your knees, spread your legs shoulder-width apart, and keep your feet firmly planted on the ground. Sink your shoulder blades so that your upper back is flat against the mat..."} At first, the avatar bent the knee correctly, however, it fails to faithfully follow the text description as time goes on. \textbf{(c)} Chest Fly: \textit{``Raise your head, keep your chest out, and tighten your abdominal muscles. Keep your palms forward and open your arms alternately at the same time..."} When comes to the word ``alternately", Guo \textit{et al.}\cite{HumanML3D} fails to capture the semantic information and comes to a standstill. ( Due to the different forms of data organization in different methods. In Guo \textit{et al.}\cite{HumanML3D}, we use skeletons to demonstrate effects. )} 

\label{fig:generation_result}
\end{figure}

\noindent \textbf{Result and Analysis.}
 Owing to the different architectural designs of algorithms, we carry over the metrics set of the original paper in each method. Results are shown in Table~\ref{tab:hag}. \textit{For category-based settings}, the FID of FLAG3D is 14.77, and the Multimod index is 6.53. In some cases, ACTOR~\cite{MathisPetrovich2021ActionConditioned3H} provides satisfactory results as shown in Figure~\ref{fig:generation_result} (a). \textit{For language-based settings}, due to the different complexity of actions in FLAG3D, the action durations are relatively long and they do not satisfy a uniform distribution. As shown in Figure~\ref{fig:generation_result} (b), TEMOS~\cite{petrovich22temos} appears to be visually plausible and context-aware at the beginning. However, it fails to follow the text description faithfully as time goes on. More structure about characterization in temporal dependencies should be designed. Guo \textit{et al.}\cite{HumanML3D} designed a list to record keywords such as body parts and movements. So that the model could focus on specific words. FLAG3D is semantically informative and has many professional descriptions. As shown in Figure~\ref{fig:generation_result} (c), Guo \textit{et al.}\cite{HumanML3D} cannot capture the information of the word ``alternately" after the execution of movement ``keep your palms forward". These cases require models of more generalization so that they could suffer from out-of-distribution descriptions. Moreover, Flag3D increased the performance of existing methods. As shown in Table \ref{tab:kit}, MDM \cite{tevet2022human} achieved a better effect in KIT \cite{MatthiasPlappert2016TheKM} after pretraining on FLAG3D. Results show that the FLAG3D dataset contains beneficial information that can transfer to other datasets.
\section{Future Works and Discussion}
\label{sec:application}
Based on the high-quality and versatile data resources of FLAG3D, some other potential directions could be further explored. We discuss some of them below:

\noindent \textbf{Visual Grounding}. 
The language instructions of FLAG3D involve the critical steps of specific body parts to accomplish an activity. Grounding these key phases with the corresponding spatial-temporal regions could better bridge the domain gap between linguistic and visual inputs.
 
\noindent \textbf{Repetitive Action Counting}. 
Counting the occurring times of the repetitive actions benefits users for fitness training~\cite{aifit,DBLP:conf/cvpr/HuDZLLG22}. While it requires more fine-grained annotations of the temporal boundary, it is desirable to explore unsupervised or semi-supervised learning methods in this direction. 

\noindent \textbf{Action Quality Assessment}.
This task aims to assess how well a fitness activity is performed and give feedback to users to avoid injury and improve the training effect. Unlike previous works ~\cite{tang2020uncertainty,xu2022finediving,parmar2017learning,shiyi}, the future effort could be devoted to FLAG3D by evaluating whether a 3D activity meets the rule described by the language instruction.

\section{Conclusion}
In this paper, we have proposed FLAG3D, a large-scale comprehensive 3D fitness activity dataset which shares the merits over previous datasets from various aspects, including highly accurate skeleton, fine-grained language description, and diverse resources. Both qualitative and quantitative experimental results have shown that FLAG3D poses new challenges for multiple tasks like cross-domain human action recognition, dynamic human mesh recovery, and language-guided human action generation. We hope the FLAG3D will promote in-depth research and more applications on fitness activity analytics for the community.

\textbf{Acknowledgments}. This work was sponsored in part by the National Natural Science Foundation of China (Grant No. 62206153, 62125603), CAAI-Huawei MindSpore Open Fund, Shenzhen Key Laboratory of next generation interactive media innovative technology (Grant No: ZDSYS20210623092001004), Young Elite Scientists Sponsorship Program by CAST (No. 2022QNRC001), and Shenzhen Stable Supporting Program (WDZC20220818112518001).

%%%%%%%%% REFERENCES
{\small
\bibliographystyle{ieee_fullname}
\bibliography{egbib}
}

\end{document}